\documentclass[a4paper,twoside]{article}

\usepackage{epsfig}
\usepackage{subcaption}
\usepackage{calc}
\usepackage{amssymb}
\usepackage{amstext}
\usepackage{amsmath}
\usepackage{amsthm}
\usepackage{multicol}
\usepackage{pslatex}
\usepackage{apalike}
\usepackage{graphicx}

\usepackage{SCITEPRESS}     % Please add other packages that you may need BEFORE the SCITEPRESS.sty package.

\begin{document}

\title{Learning Unsupervised Cross-domain Image-to-Image Translation Using a Shared Discriminator}

\author{\authorname{Rajiv Kumar\sup{1}\orcidAuthor{0000-0003-4174-8587}, Rishabh Dabral\sup{1} and G. Sivakumar\sup{1}\orcidAuthor{0000-0003-2890-6421}} \affiliation{\sup{1}CSE Department, Indian Institute of Technology Bombay, Mumbai, India} \email{\{rajiv, rdabral\}@cse.iitb.ac.in, siva@iitb.ac.in}}


\abstract{Unsupervised image-to-image translation is used to transform images from a source domain to generate images in a target domain without using source-target image pairs. Promising results have been obtained for this problem in an adversarial setting using two independent GANs and attention mechanisms. We propose a new method that uses a single shared discriminator between the two GANs, which improves the overall efficacy. We assess the qualitative and quantitative results on image transfiguration, a cross-domain translation task, in a setting where the target domain shares similar semantics to the source domain. Our results indicate that even without adding attention mechanisms, our method performs at par with attention-based methods and generates images of comparable quality.
}

\onecolumn \maketitle \normalsize \setcounter{footnote}{0} \vfill
\section{\uppercase{Introduction}}\label{sec:introduction}
\noindent Generative Adversarial Networks(GANs) \cite{Goodfellow_2014_NIPS} belong to the class of generative models \cite{Kingma2013AutoEncodingVB} widely used in various image generation and translation tasks like computer vision and image processing ~\cite{Johnson2016SR},~\cite{Wang2019SRsurvey},~\cite{Wu2017GPGAN}. While the state-of-the-art methods ~\cite{MUNIT}, ~\cite{LiuFUNIT19}, ~\cite{park2019SPADE} in image-to-image translation tasks are significantly good \cite{pix2pixHD_wang2018}, ~\cite{Mejjati2018UAIT}, ~\cite{attentionGAN} across multi-domain and cross-domain tasks, there is still room for improvement in image transfiguration tasks. Most of the image-to-image translation tasks assume the availability of source-target image pairs \cite{pix2pix_isola2017}, \cite{zhu2017BicycleGAN} or expect the source-target pairs to have rough alignment between them \cite{pix2pix_isola2017}, \cite{pix2pixHD_wang2018}. However, there are scenarios where source-target image pairs are not available or when arbitrarily selected source-target image pairs have poor alignment between them.

While most image-to-image translation tasks involve translation over the complete image, there are cases where only an object of interest needs to be translated in the source and target domain. Let's consider the case of translating images of apples to oranges or horses to zebras. In both cases, only the object of interest needs to be translated, without affecting the rest of the image or it's background. This calls for the need of attention mechanisms \cite{KastaniotisATAGAN18}, \cite{qian2017AttentiveGAN}, \cite{zhang2018SAGAN}, \cite{talreja2019AGCoGAN} to attend to the objects of interest. Contrast-GAN \cite{ContrastGAN} is a work that has used object-mask annotations to guide the translation at high-level semantic levels at the cost of extra data. However, recent works have used attention mechanisms \cite{Residual-Attention_WangJQYLZWT17}, \cite{Mejjati2018UAIT}, \cite{attentionGAN} without using any extra data or pretrained models. Moreover, very few works focus on image transfiguration, a cross-domain image-to-image translation task in an unsupervised setting without using additional networks, extra data or attention mechanisms. 

In this paper, we focus on the above problem by proposing a framework that unifies the capabilities of multiple discriminators into a shared one, which not only improves the efficacy but also works without using extra data(object masks) or attention mechanisms. Adversarial training of the network involves combining the labels of the domains from different tasks conditioned on the input image and optimizing the objectives of the networks. We believe that there has not been any previous work where a dual generator shared discriminator setup has been used for \textbf{cross-domain image-to-image translation} and we are the first to propose a novel method. We summarize the paper contribution as follows:
\begin{enumerate}
\item 
We improve the efficacy of the GANs used for unsupervised cross-domain image-to-image translation tasks by introducing a novel shared discriminator setup. We empirically demonstrate the effectiveness of our method on image transfiguration tasks and report the qualitative and quantitative results on two datasets.
\item
We conduct an ablation study to study the efficacy of the networks, training objectives and architectures keeping the dataset and other parameters constant and report the quantitative results of the study.
\end{enumerate}
%-------------------------------------------------------------------------
\section{Related work}
\noindent \textbf{Generative Adversarial Networks:} GANs are generative networks that use a trainable loss function to adapt to the differences between the data distributions of generated images and the real images. Since their inception \cite{Goodfellow_2014_NIPS} \cite{radford2015GAN}, GANs have been used in various applications from computer vision \cite{DBLP:journals/corr/MaNIPS17}, \cite{DBLP:journals/corr/VondrickNIPS16}, image-to-image translation \cite{TaigmanICLR17}, \cite{TungAIGNICCV17}, video-to-video translation \cite{Fewshotvid2vid}, \cite{wang2018vid2vid}, image super-resolution \cite{DBLP:journals/corr/LedigCVPR17}, etc. among others. We refer interested readers to read more about GANs from \cite{Creswell_2018},\cite{jabbar2020survey}, \cite{DBLP:journals/corr/Kurach18} and \cite{wang2020generative}.

\noindent
\textbf{Image-to-image translation:} Recent image-to-image translation works like pix2pix \cite{pix2pix_isola2017}, pix2pixHD \cite{pix2pixHD_wang2018} use conditional GANs to learn a mapping from source domain images to target domain images.  While some rely on paired source-target images, works like CycleGAN, DualGAN, DiscoGAN \cite{DiscoGAN_KimCKLK17} and \cite{TungAIGNICCV17}, \cite{TaigmanICLR17}, \cite{CoGAN_0001T16}, \cite{UNIT_LiuBK17}, \cite{BousmalisSDEK16} learn the mapping between the source domain and target domain without using any paired images. CoGAN \cite{CoGAN_0001T16} also learns the joint distribution of multi-domain images by sharing weights of generators and discriminators. UNIT \cite{UNIT_LiuBK17} uses a shared latent space framework built on CoGANs to learn a joint distribution of different domain images and achieves very high quality image translation results.
\par

In an adversarial setting, image-to-image translation involves generators that learn mappings to translate images from a source domain to a target domain and vice-versa. Furthermore, adversarial methods that involve GAN either share network weights \cite{CoGAN_0001T16}, \cite{talreja2019AGCoGAN} or use mechanisms \cite{DualGAN_YiZTG17}, \cite{CycleGAN2017} that involve a primal GAN and a dual GAN. A Dual-GAN (or DualGAN) \cite{DualGAN_YiZTG17} setup employs two discriminators: a primal GAN and a dual GAN, performing inverse tasks of each other. Each discriminator is trained to discriminate target domain images as positive samples and translated source domain images as negative samples. Similarly, in CycleGAN \cite{CycleGAN2017}, the primal-dual relation is regularized by a forward consistency loss and backward cycle consistency loss, which constitutes the cycle-consistency loss. This reduces the space of possible mappings by enforcing a strong relation across domains.

Conventionally, separate task-specific generators and discriminators are needed for image-to-image translation, since each network deals with a different set of real and fake images. However, StarGAN \cite{Choi2018CVPRStarGAN} achieves multi-domain image translation using a single generator by considering each domain as a set of images with a common attribute (for e.g. hair color, gender, age, etc.) and by exploiting the commonalities in the datasets. Similarly, a Dual Generator GAN($G^2GAN$) \cite{Tang2019G2GAN} consists of two task-specific generators and single discriminator focusing on multi-domain image-to-image translation. However, their optimization objective is complex, consisting of five components including color consistency loss, MS-SSIM loss and conditional identity preserving loss for preventing mode collapse. While Dual Generator GAN uses a single discriminator, the underlying task is multi-domain image translation. However, in this paper we focus on the task of cross-domain image translation using a single shared discriminator.
\section{Methodology}
\noindent
We briefly explain the problem formulation in subsection \ref{Problem Formulation}, proposed framework in subsection \ref{Proposed Framework}, image pools in subsection \ref{Image Pools}, training stages in subsection \ref{training stages} and loss functions in subsection \ref{Loss Functions} below.
\subsection{Problem Formulation}
\label{Problem Formulation}
\noindent
For the image-to-image translation problem, our goal is to learn two mapping functions, $G_{AB}:A\rightarrow B$ and  $G_{BA}:B\rightarrow A$, between domains $A$ and $B$ modelled by generators $G_{AB}$ and $G_{BA}$ respectively. We consider task-specific generators since the input distribution is different for each task in cross-domain image-to-image translation. A domain is referred to as either \textit{source} or \textit{target} domain, based on its role in the translation task. The goal of the generator $G_{AB}$ is to translate an input image $a$ from source domain $A$ to the target domain, such that the generated image $b^*$ follows the distribution of the target domain $B$, $p_{data}(B)$. Likewise, the task of generator $G_{BA}$ is to translate an image $b \in B$ to an image $a^*$ such that it follows the distribution of the target domain $A$, $p_{data}(A)$. We propose to provide adversarial supervision using a novel shared discriminator, $D_{shared}$ common to both the generators without using extra networks, masks or additional data. In this paper, we focus our method on transfiguration tasks, which requires translation of objects of interest while keeping other objects and the background same. Some transfiguration tasks include apples $\leftrightarrow$ oranges, horses $\leftrightarrow$ zebras, etc.
\subsection{Proposed Framework}
\label{Proposed Framework}
\noindent
Each translation task ($A \rightarrow B$ and $B \rightarrow A$) is mapped to a separate generator. For guided image generation, we use conditional GANs \cite{cGAN_MirzaO14} that condition using the input images. During training, each generator learns to translate its input from \textit{source} domain to the corresponding \textit{target} domain. However, our approach differs from the conventional setting, which treats the target domain samples as \textit{real} and translated images as \textit{fake}. Instead, we exploit the fact that the data distributions of the source and the target domains of one translation task are the same as that of the target and the source domains of its inverse translation task.

In our novel formulation, the proposed shared discriminator $D_{shared}$ is trained to classify the generated images into either belonging to domain $A$ or domain $B$. The translated images and random images from the two domains are conditioned on the input images to form the base for adversarial training using the shared discriminator. We hypothesize that this unification allows for \textit{domain-aware} discrimination which is crucial for tasks like transfiguration, where a specific part of the image with distinct feature sets are to be transformed. GANs are infamous for unstable training and prone to model oscillation. To stabilize the model training, we leverage the power of image pools with modifications tailored for our approach. Once the training is complete, the generator outputs are treated as final prediction and the discriminators are not needed in inference stage. 
%----------------------------------
\begin{figure}[t] \small
\centering
\includegraphics[width=0.95\linewidth]{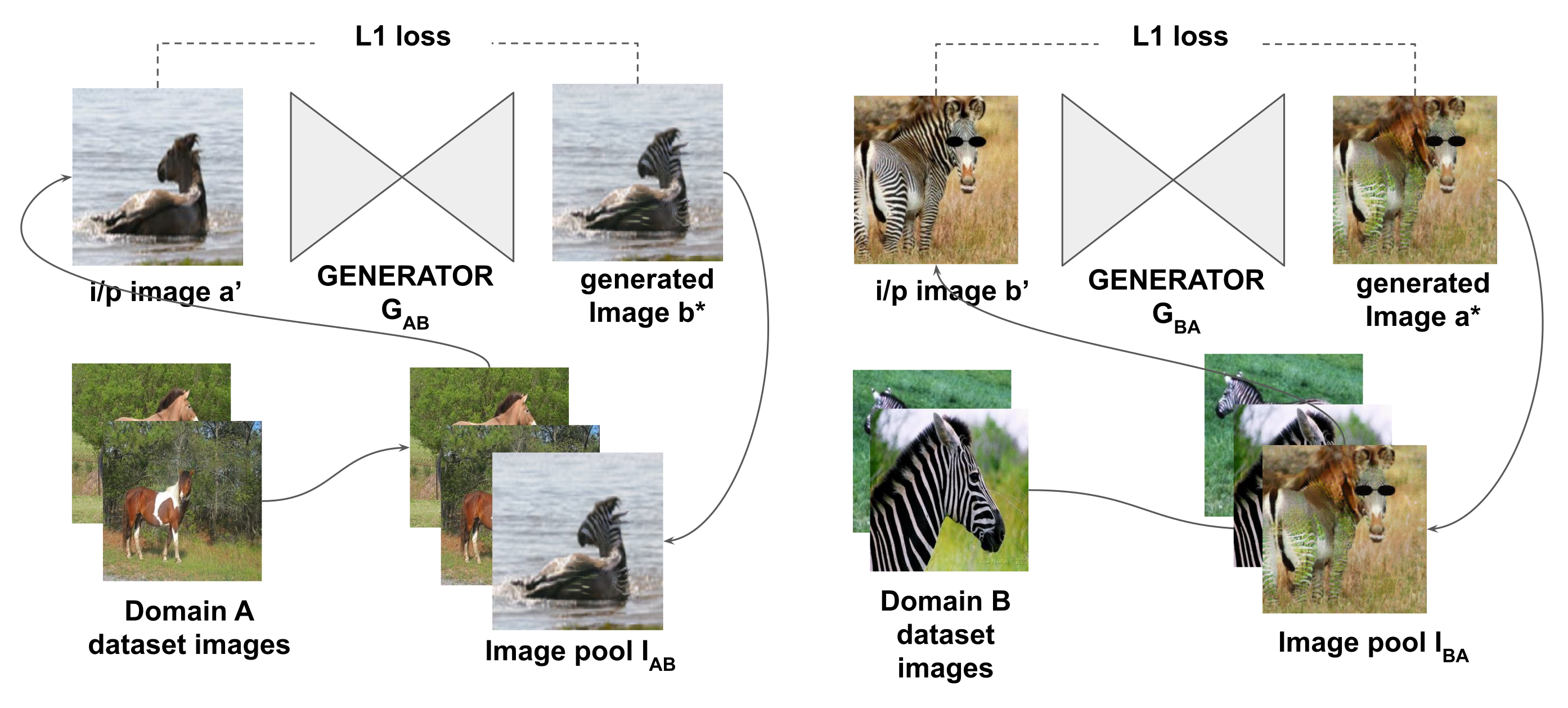}
\caption{The above image corresponds to training stage 1, with two generators and two image pools. Here the generated images are pushed to the same image pool as that of the translation task. Shared discriminator has been avoided for brevity.} 
\label{fig:Training Stage 1}
\centering
\includegraphics[width=0.95\linewidth]{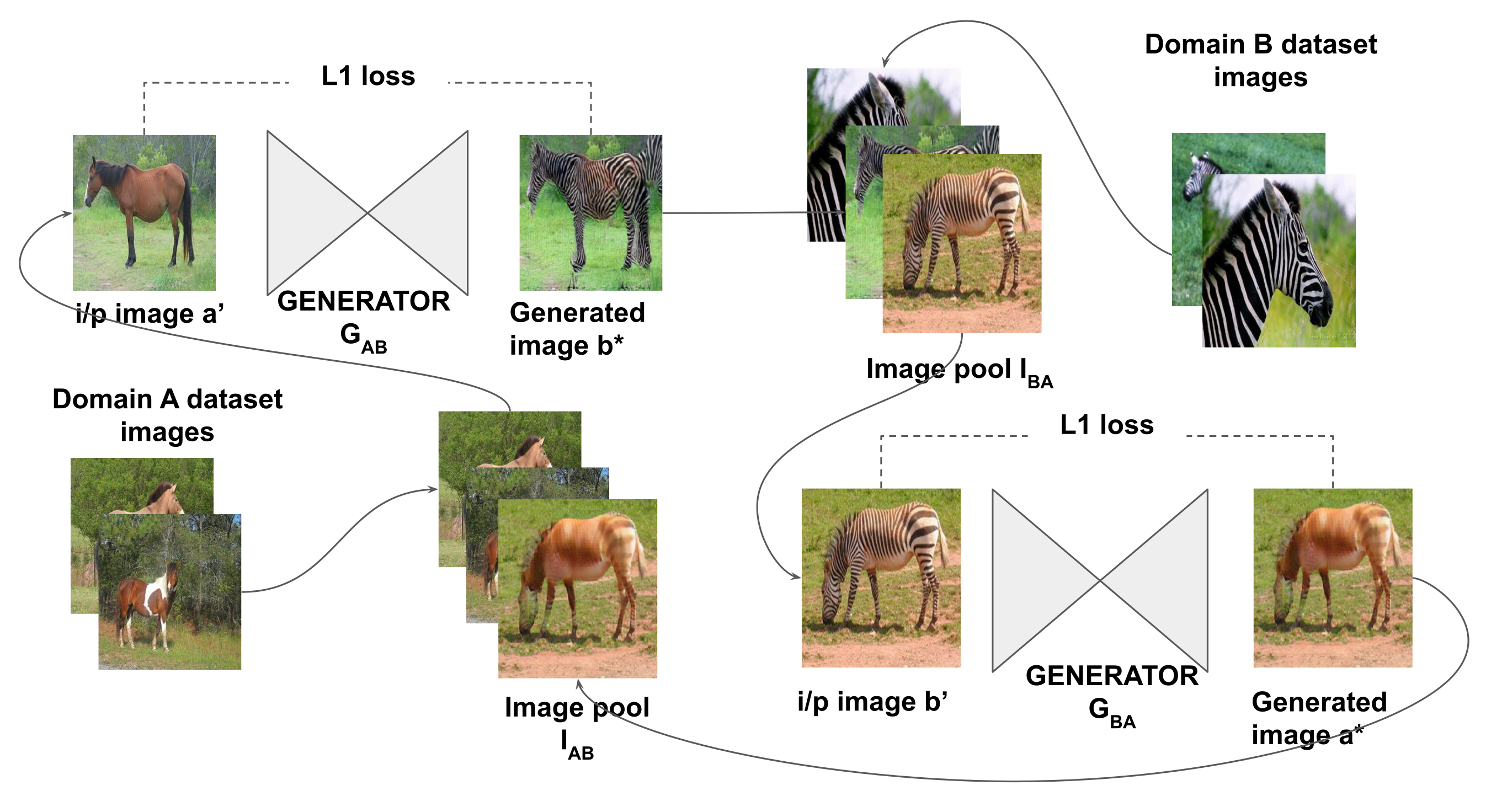}
\caption{The above image corresponds to training stage 2, with two generators and two image pools. Here the generated images are pushed to the image pool of the corresponding inverse translation task. Shared discriminator has been avoided for brevity.}
\label{fig:Training Stage 2}
\end{figure}
%----------------------------------
\subsubsection{Image Pools}
\label{Image Pools}
Generally, the generator outputs are reused in image-to-image translation techniques that involves a reconstruction loss between the source image and the reconstructed image(the resulting image after undergoing two translations, from source domain to the target domain and back to the source domain). An image pool \cite{DBLP:journals/corr/ShrivastavaPTSW16} is generally used to store a history of generated images to feed the discriminator in order to reduce the model oscillation during adversarial training. In our method, we associate an image pool to the generator of each translation task, such that the translated images can be reused as inputs to either of the generators by pushing to one image pool or the other, i.e. image pool $I_{AB}$ is associated with $G_{AB}$ and image pool $I_{BA}$ is associated with $G_{BA}$ (see Fig.~\ref{fig:Training Stage 1} and Fig.~\ref{fig:Training Stage 2}). We use this simple tweak to improve the robustness of the generators to deal with variety of input images. In some cases, we also observe performance improvements, which we discuss later in the ablation study. Since the generated images are pushed to the image pools, each image pool gets a static input set of source domain images and an evolving input set of generated images from one generator or the other, depending upon the training stage.
\subsection{Training stages:}
\label{training stages}
\noindent
We consider the image-to-image translation task $A \leftrightarrow B$ by learning the translation mapping $G_{AB}:A\rightarrow B$ and it's inverse translation mapping $G_{BA}:B\rightarrow A$. Throughout the training process, the inputs to the generators $G_{AB}$ and $G_{BA}$ are from the image pools $I_{AB} = \{a'_1, a'_2, \dots, a'_{|I_{AB}|}\}$ and $I_{BA} =  \{b'_1, b'_2, \dots, b'_{|I_{BA}|}\}$ respectively. The image pools are initialized by the images of their source domains, $A$ and $B$. The details of each training stage are given below.
\\
\noindent \textbf{Training stage 1:} If we consider the initial stages of training, the translated images appear closer in appearance to the source domain with very few target domain features. Therefore, we interleave the translated images from the generator with the source domain images using the same image pool, i.e. $I_{AB}$ would pool-in images $a$ from $A$ and $b*$ from $G_{AB}(a)$, and $I_{BA}$ would pool-in images $b$ from $B$ and $a*$ from $G_{BA}(b)$ as depicted in Fig.~\ref{fig:Training Stage 1}.
\\
\noindent \textbf{Training stage 2:} As training proceeds, the generators improve upon their translation capabilities and the generated images possess more target domain features and very few source domain features. Therefore, each generator can take the outputs of the other generator as adversarial images, in addition to their respective source domain images, i.e. $I_{AB}$ would pool-in images $a$ from $A$ and $a*$ from $G_{BA}(b)$, and $I_{BA}$ would pool-in images $b$ from $B$ and $b*$ from $G_{AB}(a)$ as depicted in Fig.~\ref{fig:Training Stage 2}. The generated images are pushed to the image pool of the inverse translation task, to mimic cyclic translations as done in some related works.
\subsection{Loss Functions}
\label{Loss Functions}
\noindent
Conventionally, a discriminator is used to distinguish between \textit{real} images from the dataset and \textit{fake} images generated by the generator. However, we avoid the usage of the terms \textit{real} images and \textit{fake} images, and use abstract binary labels \textit{True} and \textit{False} instead. We assign the same labels for a domain irrespective of the translation task or their role in the translation task, i.e. the label assigned for the source domain images in the forward translation is the same as that of the target domain images in the inverse translation task. We assign \textit{true} labels for domain B images and \textit{false} labels for domain A images.

\noindent\textbf{Discriminator loss:} 
All translated images are conditioned on the their input images when subjected to the discriminator, while optimizing the objectives of $D_{shared}$, i.e. $b^*$ is conditioned on $a'$ and $a^*$ is conditioned on $b'$.
The generated images, $a^*$ from $G_{AB}(a')$ and $b^*$ from $G_{BA}(b')$ are labelled the same labels as their source domain images $a$ and $b$ respectively, while subjecting to the discriminator.
The shared discriminator $D_{shared}$ is trained with a binary cross entropy loss $L_{D_{shared}}$.
% $A'\int I_{AB}, B'\int I_{BA}, $ and $(a^*, b^*)$,
The goal of $D_{shared}$ is to classify the generated images into either domain $A$ or domain $B$ depending upon the source domain of the translation task. In addition, we subject the shared discriminator to random domain $B$ images labelled as \textit{true} and random domain $A$ images labelled as \textit{false}. These random images are conditioned on input images $a'$ or $b'$ depending on the translation task or whether they are input to generator $G_{AB}$ or $G_{BA}$ respectively. Formally, the complete training objective of $D_{shared}$ or the discriminator loss function is given by,
\begin{align}
{L}_{D_{shared}}(G_{AB}, G_{BA}, D_{shared}, A, B, I_{AB}, I_{BA}) = \nonumber\\ {E}_{b\sim{p_{\rm data}}(b)}[log(D_{shared}(b|a'))] + \nonumber\\ 
{E}_{a\sim{p_{\rm data}}(a)}[log(1 - D_{shared}(a|a'))] + \nonumber\\
{E}_{a'\sim{p_{\rm data}}(a')}[log(1 - D_{shared}(G_{AB}(a')|a'))] + \nonumber\\ 
{E}_{a\sim{p_{\rm data}}(a)}[log(1 - D_{shared}(a|b'))] + \nonumber\\
{E}_{b\sim{p_{\rm data}}(b)}[log(D_{shared}(b|b'))] + \nonumber\\ 
{E}_{b'\sim{p_{\rm data}}(b')}[log(D_{shared}(G_{BA}(b')|b'))].
\label{Equ: Adversial discriminator 1}
\end{align}
The first three parts of Eq.~\ref{Equ: Adversial discriminator 1} are conditioned on input images $a'$ from image pool $I_{AB}$ and represent the translation $A\rightarrow B$, while the latter parts are conditioned on the input images $b'$ from image pool $I_{BA}$ and represent the translation $B\rightarrow A$.

\noindent\textbf{Generator loss:}
We enforce a reconstruction loss between the generator's input and it's output involving only one image translation, in contrast to conventional pixel reconstruction objectives that involves translations over both directions. We choose a loss function that can preserve the median values, so that the objects of interest are translated without translating other objects in the image or the background. This motivates the use of $L_1$ pixel reconstruction loss between the input and output of each generator with additional help from adversarial training. The adversarial goal of each generator is to fool the shared discriminator into identifying generated images as belonging to the target domain images, i.e. $G_{AB}$ tries to map $b^*$ as belonging to $B$ while $G_{BA}$ tries to map $a^*$ as belonging to $A$. The adversarial losses overrule the reconstruction loss over the membership score of the generated image, which results in the source images to take target domain features. We can express the full objective of $G_{\text{AB}}$ as the sum of Eq.~\ref{Equ: Generator GAN loss AtoB}, which corresponds to the adversarial loss and Eq.~\ref{Equ: L1_Generator AtoB}, which corresponds to the $L_1$ reconstruction loss. Similarly, we can express the full objective of $G_{\text{BA}}$ as the sum of Eq.~\ref{Equ: Generator GAN loss BtoA}, which corresponds to the adversarial loss and Eq.~\ref{Equ: L1_Generator BtoA}, which corresponds to the $L_1$ reconstruction loss.
\begin{align}
{L}_{{G_{\text{AB}}}}(G_{AB},D_{shared}, I_{AB}) &=  \nonumber\\
{E}_{a'\sim{p_{\rm data}}(a')}[log(D_{shared}(G_{AB}(a')))].
\label{Equ: Generator GAN loss AtoB}
\\
{L}_{pixel}^{G_{\text{AB}}}(G_{AB}, I_{AB}) &= \nonumber\\
{E}_{a'\sim{p_{\rm data}}(a')}[\Arrowvert G_{AB}(a')-a'\Arrowvert_1].
\label{Equ: L1_Generator AtoB}
\\
{L}_{G_{\text{BA}}}(G_{BA}, D_{shared}, I_{BA}) &= \nonumber\\ {E}_{b'\sim{p_{\rm data}}(b')}[log(1 - D_{shared}(G_{BA}(b')))].
\label{Equ: Generator GAN loss BtoA}
\\
{L}_{pixel}^{G_{\text{BA}}}(G_{BA},I_{BA}) &= \nonumber\\
{E}_{b'\sim{p_{\rm data}}(b')}[\Arrowvert G_{BA}(b')-b'\Arrowvert_1].
\label{Equ: L1_Generator BtoA}
\end{align}
%------------------------------------------------------------------------
\section{Implementation}
\label{Implementation}
\noindent
We trained the tasks on 128x128 size images as well as on 256x256 size images. For training, the training images were resized to 1.125 times and were randomly cropped to the required size. The batch size for all our experiments was 4. Smaller batch sizes enable training with larger image sizes. Also, the image pools could be stored in the main memory or cuda device memory. We experimented with the \textit{Adam} optimizer as well as \textit{RMSProp}, and found that \textit{Adam} gives better performance for most of our experiments. We used a learning rate of 0.0001 with the \textit{Adam} optimizer with betas of 0.5 and 0.999. We used the adversarial loss for membership score with the vanilla GAN or binary cross entropy with logit loss. We used a lambda of 10.0 for the adversarial losses and a lambda in [100.0, 200.0] for the reconstruction loss. 
%------------------------------------------------------------------------
\subsection{Architecture:}
\label{Architecture}
\noindent
We use identical network architecture for both the generators throughout an experiment. We conduct experiments with the Resnet \cite{Zhang2015Resnet} architecture as well as Unet \cite{Unet_RonnebergerFB15} architecture. While using the Unet architecture, the generator has the same number of downsampling layers and upsampling layers with a bottleneck in between and skip connections connecting the downsampling and upsampling layers. Our proposed method doesn't use noise vectors as in the pix2pix implementation\cite{pix2pix_isola2017}. Also, using dropout doesn't affect the performance of our method when implemented with the Unet architecture. In the Resnet architecture, the skip connections exist between Resnet blocks. The discriminator's architecture used in our experiments is PatchGAN \cite{CycleGAN2017}. 
%------------------------------------------------------------------------
\begin{table}[!ht] \footnotesize
\centering
\caption{Effect of network architectures \cite{Unet_RonnebergerFB15} and \cite{Zhang2015Resnet} on translation tasks horse $\leftrightarrow$ zebra and apples $\leftrightarrow$ oranges. The results are compared using FID and KID scores.}
\resizebox{0.95\linewidth}{!}{%
\begin{tabular}{|l|c|c|c|c|} \hline
\textbf{FID}  & \textbf{Horse} & \textbf{Zebra} & \textbf{Apples} & \textbf{Oranges} \\ \hline
Unet & 211.76 $\pm$ 3.65 & 119.99 $\pm$ 14.01& \textbf{164.87} $\pm$ \textbf{4.20} & 172.30 $\pm$ 2.33 \\ \hline
Resnet & \textbf{210.37} $\pm$ \textbf{5.10} & \textbf{97.47} $\pm$ \textbf{7.85} & 168.86 $\pm$ 3.20 & 172.30 $\pm$ 2.33 \\ \hline
\textbf{KID} & \textbf{Horse} & \textbf{Zebra} & \textbf{Apples} & \textbf{Oranges} \\ \hline
Unet & 0.063$\pm$ 0.002 & 0.046$\pm$0.003  & \textbf{0.051} $\pm$ \textbf{0.003} & 0.044$\pm$ 0.002\\ \hline
Resnet & \textbf{0.058 $\pm$ 0.002} & \textbf{0.030} $\pm$ \textbf{0.002} & 0.052 $\pm$ 0.002 & 0.044 $ \pm$ 0.002 \\ \hline
\end{tabular}}
\label{tab: Comparison of Network Architectures}
\end{table}
%-----------------------------------
\begin{figure}[!ht] \small
\centering
\includegraphics[width=0.95\linewidth]{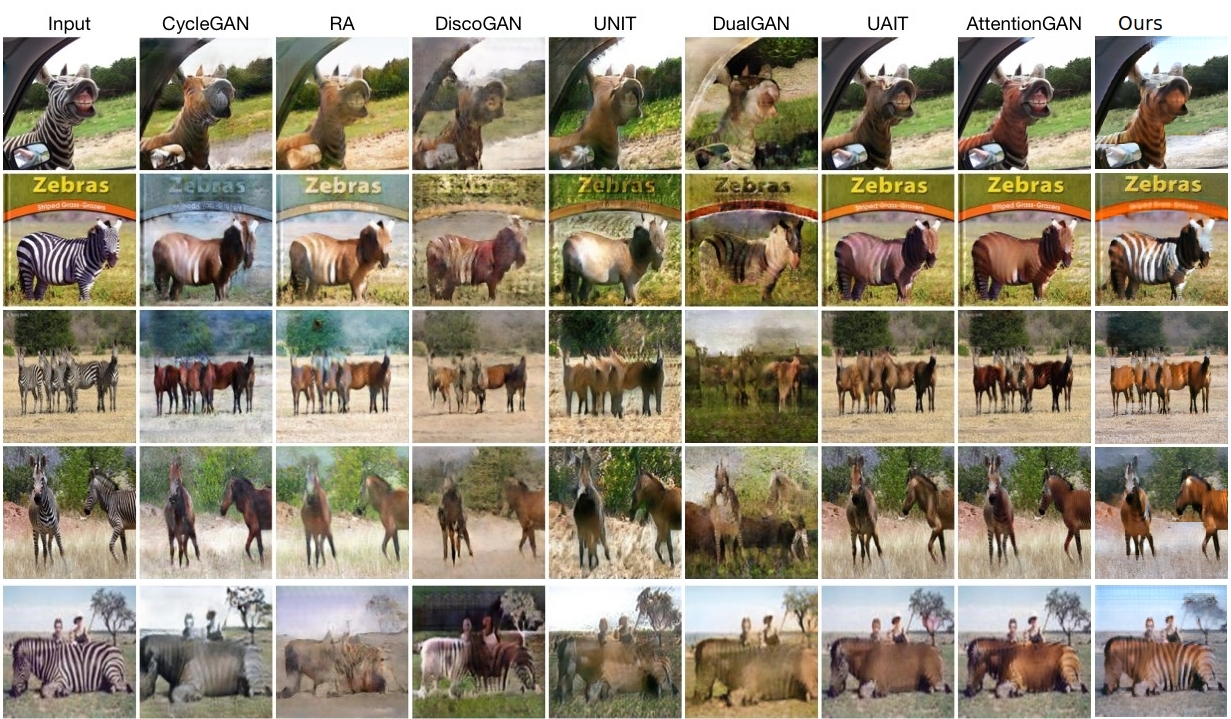}
\caption{Comparison of test images generated by different methods \cite{CycleGAN2017}, \cite{Residual-Attention_WangJQYLZWT17}, \cite{DiscoGAN_KimCKLK17}, \cite{UNIT_LiuBK17}, \cite{DualGAN_YiZTG17}, \cite{Mejjati2018UAIT}, \cite{attentionGAN} (left to right) on zebra to horse task. Leftmost column shows the input, rightmost column shows results from our method.
}
\label{img:Comparison zebra to horse}
\end{figure}
%-----------------------------------
\begin{figure}[!ht] \small
\centering
\includegraphics[width=0.95\linewidth]{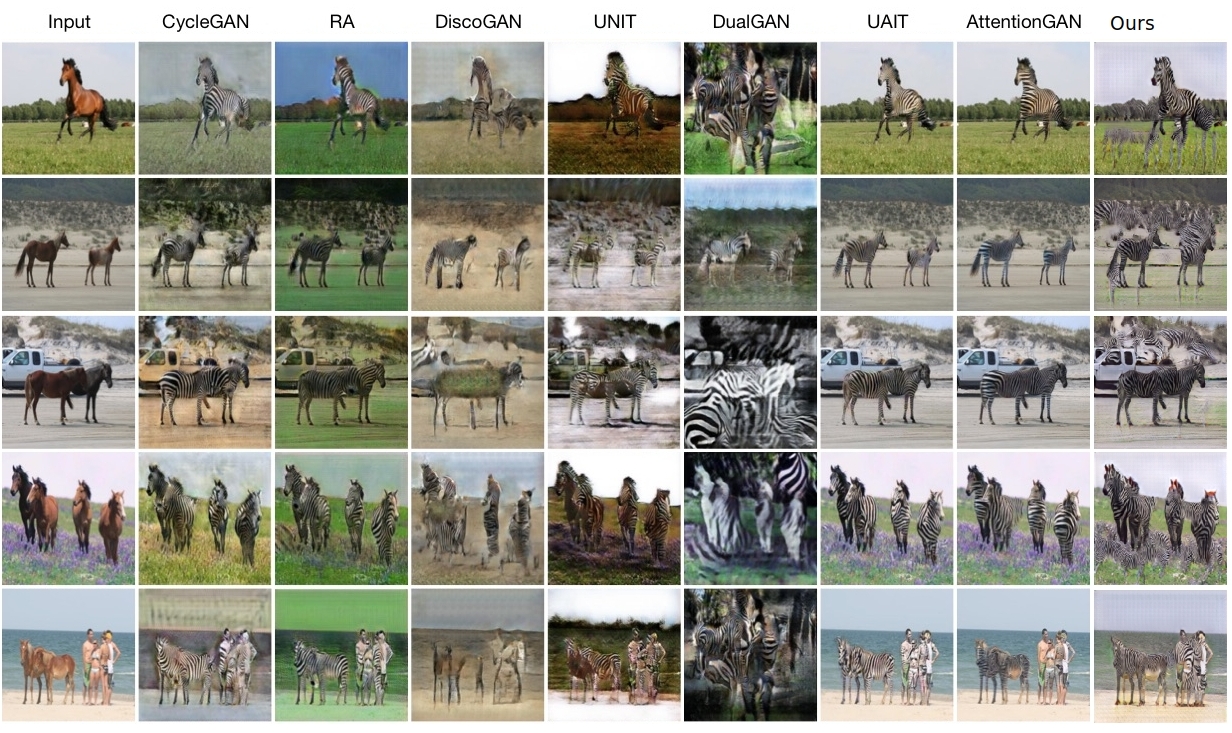}
\caption{ Comparison of test images generated by different methods \cite{CycleGAN2017}, \cite{Residual-Attention_WangJQYLZWT17}, \cite{DiscoGAN_KimCKLK17}, \cite{UNIT_LiuBK17}, \cite{DualGAN_YiZTG17}, \cite{Mejjati2018UAIT}, \cite{attentionGAN} (left to right) on horse to zebra task. Leftmost column shows the input, rightmost column shows results from our method.
}
\label{img:Comparison horse to zebra}
\end{figure}
%-----------------------------------
%------------------------------------------------------------------------
\section{Experiments and Evaluation}
\subsection{Datasets:}
\label{datasets}
\noindent
We used \textit{apples} to \textit{oranges} dataset and \textit{horse} to \textit{zebra} dataset which were originally used in CycleGAN \cite{CycleGAN2017}. These images are available from Imagenet with a training set size of each class having 939 (horse), 1177 (zebra), 996 (apple), and 1020 (orange) images.
\begin{table}[!ht] \footnotesize
\centering
\caption{FID scores between generated samples and target samples for horse to zebra translation task on methods \cite{UNIT_LiuBK17}, \cite{CycleGAN2017}, \cite{Yang2018SAT}, \cite{attentionGAN} (from top to bottom). For this metric, lower is better.}
\resizebox{0.60\linewidth}{!}{%
\begin{tabular}{|l|c|}
\hline
\textbf{Method} & Horse $\rightarrow$ Zebra \\ \hline
UNIT &  241.13 \\ \hline
CycleGAN &  109.36 \\ \hline
SAT (Before Attention) & 98.90  \\ \hline
SAT (After Attention) &  128.32 \\ \hline
AttentionGAN & \textbf{68.55} \\ \hline
Ours & 92.91 \\ \hline
\end{tabular}}
\label{tab:FID score comparison on Horse to Zebra task}
\end{table}
\subsection{Evaluation metric:}
\label{evaluation metric}
\noindent
We use the Frechet Inception Distance(FID) \cite{FID_HeuselRUNKH17} and Kernel Inception Distance(KID) \cite{KID_MMD_GAN} preferably over metrics like Inception score. For both metrics, lower scores imply similarities in features between the compared sets of images. However, both metrics are adversely affected by the presence of adversarial noise and hallucinated features in the generated images that these metrics do not correlate to the judgement by human perception. This suggests that either metrics aren't better than each other, and better scores doesn't always imply better translation results. Hence, we consider those FID and KID scores from our experiments which are positively correlated. 
%-----------------------------------------------------------------------
\begin{table}[!hb]
\centering
\caption{KID $\times$ 100 $\pm$ std. $\times$ 100 compared for different methods \cite{DiscoGAN_KimCKLK17}, \cite{Residual-Attention_WangJQYLZWT17}, \cite{DualGAN_YiZTG17}, \cite{UNIT_LiuBK17}, \cite{CycleGAN2017}, \cite{Mejjati2018UAIT}, \cite{attentionGAN}(from left to right). Abbreviations: (H)orse, (Z)ebra (A)pple, (O)range.}
\resizebox{0.95\linewidth}{!}{%
\begin{tabular}{|l|c|c|c|c|} \hline
\textbf{Method} & H $\rightarrow$ Z& Z $\rightarrow$ H & A $\rightarrow$ O & O $\rightarrow$ A \\ \hline
\textbf{DiscoGAN} & 13.68 $\pm$ 0.28 & 16.60 $\pm$ 0.50 & 18.34 $\pm$ 0.75 & 21.56 $\pm$ 0.80 \\ \hline
\textbf{RA} & 10.16 $\pm$ 0.12 & 10.97 $\pm$ 0.26 & 12.75 $\pm$ 0.49 & 13.84 $\pm$ 0.78 \\ \hline
\textbf{DualGAN} & 10.38 $\pm$ 0.31 & 12.86 $\pm$ 0.50 & 13.04 $\pm$ 0.72 & 12.42 $\pm$ 0.88 \\ \hline
\textbf{UNIT} & 11.22 $\pm$ 0.24 & 13.63 $\pm$ 0.34 & 11.68 $\pm$ 0.43 & 11.76 $\pm$ 0.51 \\ \hline
\textbf{CycleGAN} & 10.25 $\pm$ 0.25 & 11.44 $\pm$ 0.38 & 8.48 $\pm$ 0.53 & 9.82 $\pm$ 0.51 \\ \hline
\textbf{UAIT} & 6.93 $\pm$ 0.27  & 8.87 $\pm$ 0.26 & 6.44 $\pm$ 0.69 & 5.32 $\pm$ 0.48\\ \hline
\textbf{AttentionGAN} & \textbf{2.03 $\pm$ 0.64} & 6.48 $\pm$ 0.51 & 10.03 $\pm$ 0.66 & \textbf{4.38 $\pm$ 0.42} \\ \hline
Ours & 3.00 $\pm$ 0.20 & \textbf{5.80 $\pm$0.20}  & \textbf{4.40 $\pm$ 0.20} & 5.10 $\pm$ 0.30 \\ \hline
\end{tabular}}
\label{tab:KID score comparison on various methods}
\end{table}
\subsection{Experiments}
\begin{figure}[!ht] \small
\centering
\includegraphics[width=0.95\linewidth]{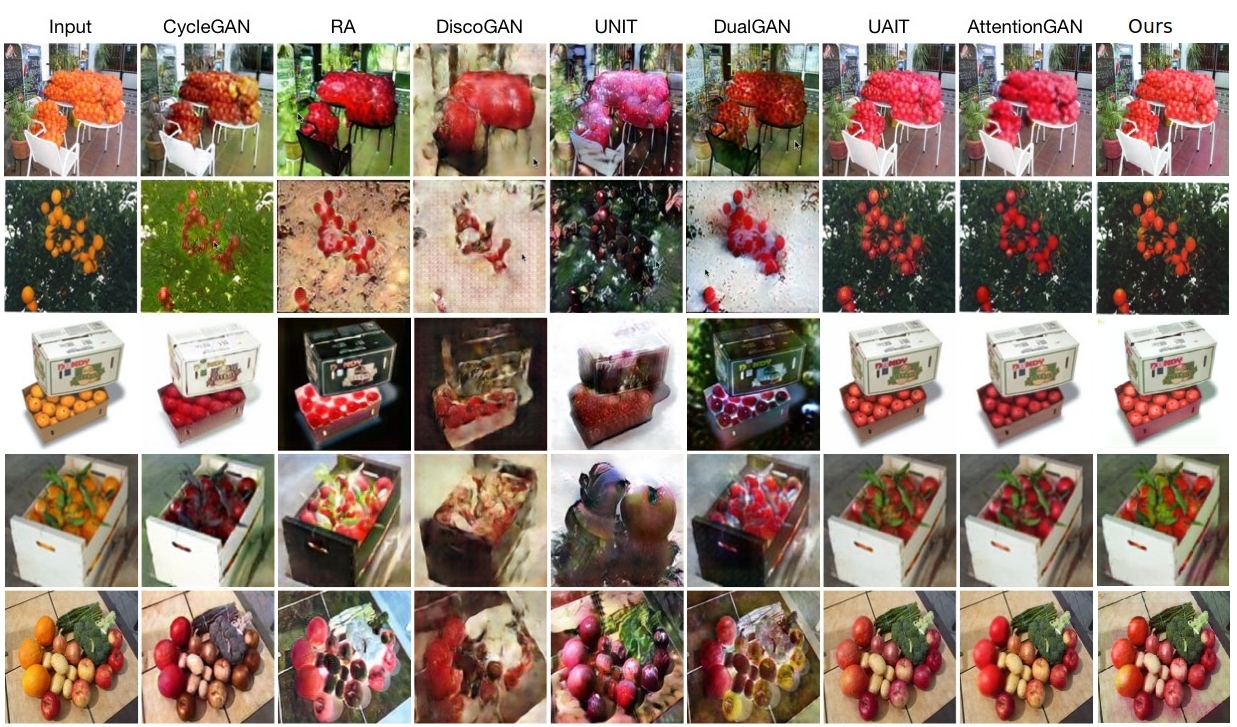}
\caption{Comparison of test images generated by different methods \cite{CycleGAN2017}, \cite{Residual-Attention_WangJQYLZWT17}, \cite{DiscoGAN_KimCKLK17}, \cite{UNIT_LiuBK17}, \cite{DualGAN_YiZTG17}, \cite{Mejjati2018UAIT}, \cite{attentionGAN} (left to right) on oranges to apples task. Leftmost column shows the input, rightmost column shows results from our method.
}
\label{img:Comparison oranges to apples}
\end{figure}

\begin{figure}[!ht] \small
\centering
\includegraphics[width=0.95\linewidth]{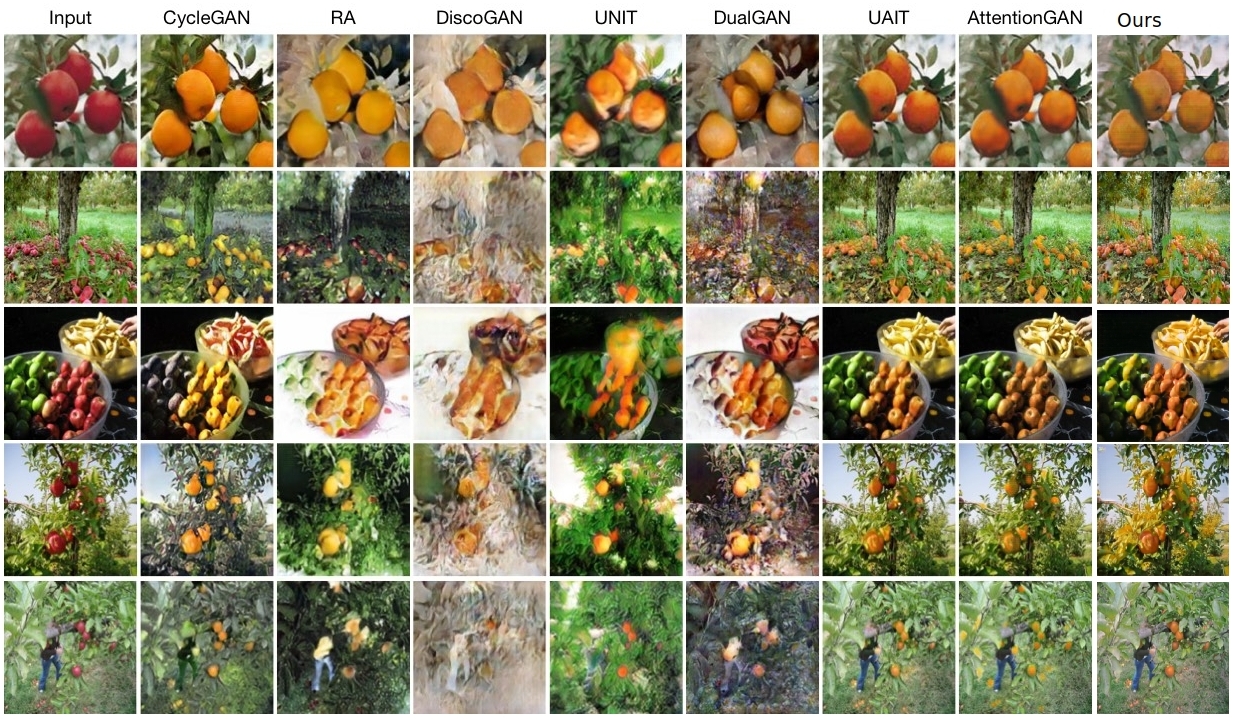}
\caption{Comparison of test images generated by different methods \cite{CycleGAN2017}, \cite{Residual-Attention_WangJQYLZWT17}, \cite{DiscoGAN_KimCKLK17}, \cite{UNIT_LiuBK17}, \cite{DualGAN_YiZTG17}, \cite{Mejjati2018UAIT}, \cite{attentionGAN} (left to right) on apples to oranges task. Leftmost column shows the input, rightmost column shows results from our method.
}
\label{img:Comparison apples to oranges}
\end{figure}

\noindent
We compute the KID score over 100 iterations and return its mean, while the FID scores are computed over 10 iterations and the mean value is returned. We compute the KID scores and FID scores on the test data using the generator models from the same checkpoint. We trained the tasks on 128x128 size images as well as on 256x256 size images and tested both category of models on 256x256 test images. We refer \cite{attentionGAN} to compile the experimental results in the qualitative comparisons and metric scores in Table~\ref{tab:KID score comparison on various methods} and \ref{tab:FID score comparison on Horse to Zebra task}. We report the performance comparison of different architectures on translation tasks \textit{horse} $\leftrightarrow$ \textit{zebra} and \textit{apples} $\leftrightarrow$ \textit{oranges}, measured in FID and KID scores in Table~\ref{tab: Comparison of Network Architectures}. We report the FID scores on horse $\rightarrow$ zebra translation task in Table~\ref{tab:FID score comparison on Horse to Zebra task}. KID scores are compared over the horses $\leftrightarrow$ zebras task and apples $\leftrightarrow$ oranges task in Table~\ref{tab:KID score comparison on various methods}. The results of qualitative comparisons includes the comparison of translated images from \textit{zebras} $\rightarrow$ \textit{horses} task in Fig.~\ref{img:Comparison zebra to horse}, \textit{horses} $\rightarrow$ \textit{zebras} task in Fig.~\ref{img:Comparison horse to zebra},
\textit{oranges} $\rightarrow$ \textit{apples} task in Fig.~\ref{img:Comparison oranges to apples}, and \textit{apples} $\rightarrow$ \textit{oranges} task in Fig.~\ref{img:Comparison apples to oranges}.
%------------------------------------------------------------------------
%-----------------------------------
\section{Results and Discussion}
\noindent
The results from Table~\ref{tab: Comparison of Network Architectures} suggest that there is a slight gain in the performance by using dropout with the Resnet network architecture for the horse $\rightarrow$ zebra task, while the same is more or less not true for apples $\rightarrow$ oranges task. We hypothesize that this observation could be due to the simplicity of the apples $\rightarrow$ oranges task while the former task is more complex. The results from Table~\ref{tab:KID score comparison on various methods} and Table~\ref{tab:FID score comparison on Horse to Zebra task} suggest that our method is at par to existing image translation methods where some related methods have an upper hand due to the underlying attention mechanisms.

On comparing the qualitative results for \textit{zebras} $\rightarrow$ \textit{horses} in Fig.\ref{img:Comparison zebra to horse}, in the second row we can notice that the text color and the background are preserved only in the translated images from UAIT~\cite{Mejjati2018UAIT}, AttentionGAN~\cite{attentionGAN} and our method. Also, our method has comparable results to \cite{Mejjati2018UAIT} and \cite{attentionGAN}, which uses attention mechanisms. While CycleGAN does a great job in translating all zebra images to horse images, the background color and tint are affected in some of the images and is severe than ours. 

On comparing the qualitative results for \textit{horses} $\rightarrow$ \textit{zebras} in Fig.~\ref{img:Comparison horse to zebra}, we notice that residual attention based method \cite{Residual-Attention_WangJQYLZWT17} generates convincing translation results, while there is a green tint in all the translated images which makes it unfavourable. Similarly, UNIT \cite{UNIT_LiuBK17} also has artifacts in the background that makes the appearance just acceptable. Our method falls behind UAIT \cite{Mejjati2018UAIT} and AttentionGAN results \cite{attentionGAN} but appears better than CycleGAN results, which has undesirable background tint in many images. Note that the translation quality drastically dropped for DualGAN \cite{DualGAN_YiZTG17}, with slightly better results from DiscoGAN \cite{DiscoGAN_KimCKLK17}.

On comparing the qualitative results for \textit{oranges} $\rightarrow$ \textit{apples} in Fig.~\ref{img:Comparison oranges to apples}, we notice that our translation results are at par with UAIT \cite{Mejjati2018UAIT} and AttentionGAN \cite{attentionGAN}, which uses attention mechanisms. CycleGAN also follows our results except that it fails in some of the images with unwanted background translations. DualGAN, DiscoGAN, UNIT and residual attention \cite{Residual-Attention_WangJQYLZWT17} fails on a task much easier than the \textit{horses} $\leftrightarrow$ \textit{zebras} task.

On comparing the qualitative results for \textit{apples} $\rightarrow$ \textit{oranges} in Fig.~\ref{img:Comparison apples to oranges}, we notice that our method consistently keeps the quality upto the mark of attention guided methods \cite{Mejjati2018UAIT}, \cite{attentionGAN}. While CycleGAN is able to translate convincingly, it is affected by a strong tint in some of the images. The translation results from UNIT and residual attention method appear similar to that of DualGAN and DiscoGAN, despite the use of attention mechanisms.
\par
\begin{table}[!hb] \small
\centering
\caption{Ablation study results on baseline, $\text{variant}_1$, $\text{variant}_2$, $\text{variant}_3$ and $\text{variant}_4$ (from left to right) evaluated using FID scores. The study included training on 128x128 size images and testing on 256x256 size images.}
\resizebox{0.95\linewidth}{!}{%
\begin{tabular}{|l|c|c|c|c|c|} 
\hline
\textbf{FID} & $D_\text{shared}$ & $D_\text{shared}1$  & No Image pool & No stage-1 & No Stage-2\\ \hline
\textbf{Horse} & \textbf{207.93} $\pm$ \textbf{6.26} & 218.74 $\pm$ 4.69  & 216.10 $\pm$ 7.659 & 221.04 $\pm$ 5.005 & 224.02 $\pm$ 6.056 \\ \hline
\textbf{Zebra} & \textbf{92.91} $\pm$ \textbf{6.58} & 100.90 $\pm$ 7.495  & 136.63 $\pm$ 10.444 & 139.77 $\pm$ 10.643 &  119.39 $\pm$ 7.025 \\ \hline
\end{tabular}}
\label{tab:Ablation study using FID metric on 128x128 images}
\end{table}
\begin{table}[!hb] \small
\centering
\caption{Ablation study results on baseline, $\text{variant}_1$, $\text{variant}_2$, $\text{variant}_3$ and $\text{variant}_4$ (from left to right) evaluated using KID scores. The study included training on 128x128 size images and testing on 256x256 size images.}
\resizebox{0.95\linewidth}{!}{%
\begin{tabular}{|l|c|c|c|c|c|}
\hline
\textbf{KID} & $D_\text{shared}$ & $D_\text{shared}1$  & No Image pool & No stage-1 & No Stage-2\\ \hline
\textbf{Horse} & \textbf{0.065} $\pm$ \textbf{0.003} & 0.084 $\pm$ 0.002  & 0.088 $\pm$ 0.002 & 0.085 $\pm$  0.002 & 0.107 $\pm$ 0.002\\ \hline
\textbf{Zebra} & \textbf{0.036} $\pm$ \textbf{0.002} & 0.047 $\pm$ 0.003  & 0.063 $\pm$ 0.003 & 0.067 $\pm$ 0.003 & 0.050 $\pm$ 0.002\\ \hline
\end{tabular}}
\label{tab:Ablation study using KID metric on 128x128 images}
\end{table}

\noindent \textbf{Ablation study:} We perform an ablation study to isolate the effects and understand the effectiveness of various components of our method using FID and KID metric over \textit{horse} $\leftrightarrow$ \textit{zebra} task comparing the baseline to different variants. We consider the original shared discriminator setup, $D_{shared}$ to be the baseline for comparing the variants. The quantitative results of the ablation study are available in Table~\ref{tab:Ablation study using FID metric on 128x128 images} and \ref{tab:Ablation study using KID metric on 128x128 images} for images trained on 128x128 size images and tested on 256x256 size images and in Table~\ref{tab:Ablation study using FID metric on 256x256 images} and \ref{tab:Ablation study using KID metric on 256x256 images} for images both trained and tested on 256x256 size images. 

\begin{table}[!hb] \small
\centering
\caption{Ablation study results on baseline, $\text{variant}_1$, $\text{variant}_2$, $\text{variant}_3$ and $\text{variant}_4$ (from left to right) trained and tested on 256x256 sizes and evaluated using FID scores.}
\resizebox{0.95\linewidth}{!}{%
\begin{tabular}{|l|c|c|c|c|c|} 
\hline
\textbf{FID} & $D_\text{shared}$ & $D_\text{shared}1$  & No Image pool & No stage-1 & No Stage-2\\ \hline
\textbf{Horse} & \textbf{212.81} $\pm$ \textbf{4.835} &  221.66 $\pm$ 6.185  & 213.64 $\pm$ 4.357 & 216.28 $\pm$  4.884 & 217.67 $\pm$ 6.864 \\ \hline
\textbf{Zebra} & \textbf{92.72} $\pm$ \textbf{9.915} & 148.95 $\pm$ 4.470  & 96.16 $\pm$ 5.251 & 118.63 $\pm$  9.380 & 113.30 $\pm$ 11.212\\ \hline
\end{tabular}}
\label{tab:Ablation study using FID metric on 256x256 images}
\end{table}
\begin{table}[!hb] \small
\centering
\caption{Ablation study results on baseline, $\text{variant}_1$, $\text{variant}_2$, $\text{variant}_3$ and $\text{variant}_4$ (from left to right) trained and tested on 256x256 sizes and evaluated using KID scores.}
\resizebox{0.95\linewidth}{!}{%
\begin{tabular}{|l|c|c|c|c|c|}
\hline
\textbf{KID} & $D_\text{shared}$ & $D_\text{shared}1$ & No Image pool & No stage-1 & No Stage-2 \\ \hline
\textbf{Horse} & \textbf{0.069} $\pm$ \textbf{0.002} & 0.090 $\pm$ 0.002 & 0.070 $\pm$ 0.002 &  0.072 $\pm$ 0.002 & 0.077 $\pm$ 0.002 \\ \hline
\textbf{Zebra} & \textbf{0.030} $\pm$ \textbf{0.002} & 0.076$\pm$ 0.004  & 0.036 $\pm$ 0.003 & 0.047 $\pm$ 0.003 & 0.045 $\pm$ 0.003 \\ \hline
\end{tabular}}
\label{tab:Ablation study using KID metric on 256x256 images}
\end{table}

First, we modify the objective of the shared discriminator in a variant $D_{shared_{1}}$ to see if all six components are really necessary. Out of the six components of the shared discriminator objective, four of them involves random source or target domain images conditioned on either $a'$ or $b'$. It may seem logical to remove two random image components of one translation task or the other to make the shared discriminator objective compact, since they differ only in the conditioned part, i.e. $a'$ or $b'$. To verify that, we deal with each domain only once and as target domain in the variant $D_{shared_{1}}$. The source domain images conditioned on the input images are not subjected to the shared discriminator and avoided in the objective assuming that the same domain images as target domain and labels will suffice. In other words, for the image translation task $A \rightarrow B$, we consider only random target domain images $b$ from $B$ with \textit{true} labels conditioned on the input images $a'$ to the shared discriminator. Analogously, for the image translation task $B \rightarrow A$, we consider only random target domain images $a$ from $A$, conditioned on the input images $b'$ with \textit{false} labels to the shared discriminator. The generator's goal and objectives are unaltered in this variant. The results in Table~\ref{tab:Ablation study using FID metric on 128x128 images}, \ref{tab:Ablation study using KID metric on 128x128 images}, \ref{tab:Ablation study using FID metric on 256x256 images} and \ref{tab:Ablation study using KID metric on 256x256 images} from the ablation study for $D_{shared_{1}}$(or \textit{$\text{variant}_1$}) indicate that irrespective of the image sizes used for training, the performance of the shared discriminator setup drops on removing the components of $D_{shared}$'s objectives. Both FID and KID values have gone higher for $D_{shared_{1}}$, which is not desirable for good image translation results.

The second variant that we consider is a shared discriminator setup without the image pool, i.e. the translated images are not reused as inputs to any of the generators. While the results in Table~\ref{tab:Ablation study using FID metric on 256x256 images} and \ref{tab:Ablation study using KID metric on 256x256 images} suggest that using image pool doesn't improve the performance of our method, the results in Table~\ref{tab:Ablation study using FID metric on 128x128 images} and \ref{tab:Ablation study using KID metric on 128x128 images} suggest that there is considerable drop in performance when the image pool is not used while training on smaller images and testing the model on larger images. We hypothesize that the generators become more robust when trained with additional translated images with the help of image pools.

The third variant that we consider is a shared discriminator setup without the training stage-1, i.e. the translated images are pushed to the image pool of the inverse translation task, throughout the training process. Similarly, the fourth variant that we consider is a shared discriminator setup without the training stage-2, i.e. the translated images are pushed back to the image pool of the same translation task throughout the training process. The results in Table~\ref{tab:Ablation study using FID metric on 256x256 images} suggest that FID values are not really affected for \textit{$\text{variant}_3$} and \textit{$\text{variant}_4$}, while
Table~\ref{tab:Ablation study using KID metric on 256x256 images} suggests that the KID values increase (or performance drops) for \textit{$\text{variant}_3$} and \textit{$\text{variant}_4$} when either training stage-1 or stage-2 is used throughout the training process. Similarly, the results from Table~\ref{tab:Ablation study using FID metric on 128x128 images} and \ref{tab:Ablation study using KID metric on 128x128 images} for \textit{$\text{variant}_3$} and \textit{$\text{variant}_4$} indicates that the performance drops on using only one of the training stages. We hypothesize that simply reusing translated images with an image pool doesn't improve the performance and can result in a drop in performance.
%-----------------------------------------------------------------------
\section{Summary}
\label{Summary}
\noindent
In this paper, we propose a framework for image transfiguration, a cross-domain image-to-image translation task improving the efficacy using a shared discriminator in an unsupervised setting. We also introduce a novel application of image pools to keep the generators more robust in the process. The qualitative and quantitative results, using metrics like FID and KID, suggest that our method, even without using masks or attention mechanisms, is at par with attention-based methods. For particular tasks, where the source domain shares similar semantics with the target domain, our method performs better than previous methods. Also, we observe that metrics like KID and FID are insufficient to evaluate the quality of translated images. They are also vulnerable to adversarial noise and hallucinated features, hampering a fair comparison of image translation methods. Future work could use attention mechanisms to further improve the results and better comparison metrics that correlate to human perception.
%------------------------------------------------------------------------
\bibliographystyle{apalike}
{\small
\bibliography{final}}
%\section*{\uppercase{Appendix}}
\end{document}